\documentclass[letterpaper, 10 pt, conference]{ieeeconf}  

\IEEEoverridecommandlockouts                              

\usepackage{graphics} 
\usepackage{amsmath} 
\usepackage{amssymb}  

\usepackage[ruled]{algorithm2e}
\usepackage{subcaption}
\usepackage{nicefrac}
\usepackage{pgfplots}
\usepackage{makecell}
\usepackage{multirow}
\usepackage{url}
\usepackage{float}

\makeatletter
\newsavebox\myboxA
\newsavebox\myboxB
\newlength\mylenA
\newcommand*\xoverline[2][0.75]{%
	\sbox{\myboxA}{$\m@th#2$}%
	\setbox\myboxB\null
	\ht\myboxB=\ht\myboxA%
	\dp\myboxB=\dp\myboxA%
	\wd\myboxB=#1\wd\myboxA
	\sbox\myboxB{$\m@th\overline{\copy\myboxB}$}
	\setlength\mylenA{\the\wd\myboxA}
	\addtolength\mylenA{-\the\wd\myboxB}%
	\ifdim\wd\myboxB<\wd\myboxA%
	\rlap{\hskip 0.5\mylenA\usebox\myboxB}{\usebox\myboxA}%
	\else
	\hskip -0.5\mylenA\rlap{\usebox\myboxA}{\hskip 0.5\mylenA\usebox\myboxB}%
	\fi}
\makeatother

\usetikzlibrary{positioning}

\title{\LARGE \bf
	Controlling an Autonomous Vehicle with Deep Reinforcement Learning*
}

\author{Andreas Folkers$^{1}$ \quad Matthias Rick$^{1}$ \quad Christof B\"uskens$^{1}$
	\thanks{*The authors would like to thank \emph{Deutsches Zentrum f\"ur Luft- und Raumfahrt (DLR) Raumfahrtmanagement, Navigation} in Bonn-Oberkassel for providing the research vehicle [grant number 50NA1615].}
	\thanks{$^{1}$WG Optimization and Optimal Control, Center for Industrial Mathematics, University of Bremen, Germany
		{\tt\small afolkers@uni-bremen.de}}
}

\newcommand{\A}{\mathcal{A}}
\renewcommand{\S}{\mathcal{S}}
\newcommand{\Ps}{\mathcal{P}}

\newcommand{\E}{\mathbb{E}}

\newcommand{\pit}{ {\pi_\theta} }

\newcommand{\pito}{ {\pi_{\theta_0}} }

\newcommand{\R}{\mathbb{R}}
\newcommand{\M}{\mathcal{M}}

\renewcommand{\O}{\mathcal{O}}
\newcommand{\Z}{\mathcal{Z}}

\DeclareMathOperator*{\clip}{clip}

\definecolor{b1}{rgb}{0.0, 0.0, 0.6}
\definecolor{b2}{rgb}{0.6, 0.6, 0.9}
\definecolor{b3}{rgb}{0.6, 0.3, 1}
\definecolor{k1}{rgb}{0.1, 0.1, 0.1}
\definecolor{k2}{rgb}{0.5,0.5,0.5}
\definecolor{r1}{rgb}{0.6, 0.0, 0.0}
\definecolor{r2}{rgb}{0.9, 0.6, 0.6}
\definecolor{o1}{rgb}{0.8,0.5,0}
\definecolor{o2}{rgb}{0.5,0.35,0}
\definecolor{g1}{rgb}{0.2,0.8,0.2}
\definecolor{g2}{rgb}{0.1,0.6,0.1}
\definecolor{m1}{rgb}{0.8,0.2,0.8}
\definecolor{c1}{rgb}{0.2,0.8,0.8}

\newcommand{\figref}[1]{Fig.\,\ref{#1}}
\newcommand{\secref}[1]{Sect.\,\ref{#1}}

\begin{document}

\maketitle
\thispagestyle{empty}
\pagestyle{empty}

\begin{abstract}

We present a control approach for autonomous vehicles based on deep reinforcement learning. A neural network agent is trained to map its estimated state to acceleration and steering commands given the objective of reaching a specific target state while considering detected obstacles. Learning is performed using state-of-the-art proximal policy optimization in combination with a simulated environment. Training from scratch takes five to nine hours. The resulting agent is evaluated within simulation and subsequently applied to control a full-size research vehicle. For this, the autonomous exploration of a parking lot is considered, including turning maneuvers and obstacle avoidance. Altogether, this work is among the first examples to successfully apply deep reinforcement learning to a real vehicle.

\end{abstract}

\section{INTRODUCTION}
Self driving cars have the potential to sustainably change modern societies which are heavily based on 
\textit{mobility}. The benefits of such a technology range from self-providing car sharing to platooning approaches, which ultimately yield a much more effective usage of vehicles and roads \cite{maurer2016}. In recent years, great progress has been made in the development of these systems, with a major factor being the results achieved through deep learning methods. One example is the neural network \textsc{PilotNet}, which was trained to steer a vehicle solely based on camera images \cite{pilotnet, understanding_pilotnet}. 

More classic methods divide the processing of sensor data and the calculation of vehicle controls into separate tasks. The latter can be achieved by various model-based control approaches, one method being the linear quadratic controller \cite{modern_control_engineering}, also known as Riccati controller. It minimizes a quadratic objective function in the state deviation and control energy while taking into account a linear model of the underlying system. There are various examples for the application of this technique to autonomous driving \cite{riccati_autonomous_driving,lqr_car_like_robot,lqr_vehicle_path_tracking}.

While a Riccati-controller is comparably fast, however, it does not directly allow the consideration of constraints such as obstacles or more advanced objective functions within the optimization. Such requirements are met by a general nonlinear model predictive control (MPC) approach based on solving an optimal control problem in every time step. Although the calculations required are considerably more complex, such methods were successfully implemented for autonomous vehicles, 
e.\,g., \cite{mpc_low_complexity} or most recently \cite{Sommer2018icatt} utilizing efficient solvers such as \textsc{TransWORHP} \cite{transworhp} based on \textsc{WORHP} \cite{worhp}.

A combination of the advantages of both, the speed of the Riccati controller and the generality of MPC, can be achieved by finding a function that maps state values to control variables, e.\,g., by training a deep neural network. Such a model could, for example, be learned supervised, as done for \textsc{PilotNet}, or by reinforcement learning.
The latter in particular led to excellent results in the training of such agents 
for controlling real-world systems such as robots \cite{rl_robots} or helicopters \cite{rl_helicopter}.

Recent work also shows promising applications of reinforcement learning for autonomous driving by making strategic decisions \cite{rl_intersections, rl_highway} or by the computation of control commands \cite{rl_driving_framework, 
	cont_qn, a3c}. 	
Although these results show the success of this approach in simulated environments, there are very few examples of evaluations on real vehicles.
One of them was presented by the \textsc{Wayve} research team where a policy for lane following is learned based on corresponding camera images. 
Training is done onboard, with the only feedback for improvement coming from the intervention of a safety driver. While this method works when training is carried out without the proximity of real obstacles and at low speeds, it may be difficult to implement this approach for more general situations. 

In this work, we show how to realize the autonomous exploration of a parking lot based on deep reinforcement learning. In particular, this setting is much more challenging than simple lane following due to sharp turning maneuvers and road constrictions caused by obstacles. To this end, we describe how a policy is trained to compute sophisticated control commands which depend on an estimate of the current vehicle state. This is done by designing an appropriate Markov decision process and a corresponding proximal policy optimization \cite{ppo} learning algorithm. For that purpose a simulated environment is used for data generation. Performance of the resulting \emph{deep controller} is evaluated in both, simulation and real-world experiments. To the best of our knowledge, this work extends the state of the art results for successfully driving an autonomous vehicle by a deep reinforcement learning policy.

\section{CONTINUOUS DEEP REINFORCEMENT LEARNING}
\label{sec:deep_reinforcement_learning}
In recent years, various methods from classical reinforcement learning \cite{rl_sutton_barto} have been combined with neural networks and their optimization through backpropagation \cite{backprob_and_momentum}, leading to algorithms such as Deep Q-Learning \cite{qn, double_qn, dueling_qn}, and several actor-critic-methods \cite{a3c,acer,trpo}. In particular, the latter class of algorithms allows agents to be trained with continuous action spaces, which is crucial for their application as controller on a real-world system like an autonomous vehicle. Our training procedure is based on the proximal policy optimization (PPO) algorithm \cite{ppo}. For this, an infinite Markov decision process (MDP) is considered, defined by the six-tuple ${(\S, \A, \Ps_{sa}^{s'},\Ps_0, r,\gamma)}$, with $\S$ and $\A$ being the continuous and bounded spaces of States and Actions, respectively. The probability density function $\Ps_{sa}^{s'}$ characterizes the transition from state $s\in \S$ to $s'\in \S$ given action $a\in\A$ while $\Ps_0$ incorporates a separate distribution of possible start states. The reward function is denoted by $r:\S\times\S\times\A\rightarrow\R$ and $\gamma$ is the discount factor.

With these definitions the goal of a reinforcement learning algorithm is to find an optimal policy, $\pi:\S \times \A\rightarrow[0,1]$ representing the probability density of the agent's action, given a certain state. Optimality is specified in relation to the expected discounted return
\begin{equation}
\begin{array}{l}
\label{eq:eta}
\eta(\pi) := \E_{s_0, a_0,\ldots}\left\{\sum\limits_{t=0}^{\infty}\gamma^t r_{t+1}\right\},\\
s_0 \sim \Ps_0(s_0), \quad a_t \sim \pi(a_t | s_t), \quad s_{t+1} \sim \Ps_{s_ta_t}^{s_{t+1}},
\end{array}
\end{equation}
with $r_{t+1} := r(s_{t+1}, s_t, a_t)$. While in the actor-critic setting the policy $\pi$ is identified with the actor, the state value function
\begin{equation*}
\begin{array}{c}
V^\pi(s_t) := \E_{a_t, s_{t+1}} \left\{\sum\limits_{k=0}^{\infty}\gamma^k r_{t+k+1}\right\}
\end{array}
\end{equation*}
will behave as the critic which is responsible to evaluate the policy's actions. Correspondingly, the state action value function and the advantage are given by
\begin{equation*}
\begin{array}{l}
Q^\pi(s_t, a_t) := \E_{s_{t+1}, a_{t+1}} \left\{\sum\limits_{k=0}^{\infty}\gamma^k r_{t+k+1}\right\} \quad \text{and} \\
A^\pi(s_t, a_t) := Q^\pi(s_t, a_t) - V^\pi(s_t).
\end{array}
\end{equation*}
To find a (nearly) optimal policy in continuous state and action spaces, it proved helpful to use neural networks as function approximators, which leads to parameterized $\pit$ and $V^\theta$. Altogether, a deep actor-critic training algorithm has to find parameters $\theta$ to both, maximize the true target \eqref{eq:eta} and approximate the corresponding state value function $V^\theta$. 

Regarding the former, proximal policy optimization is based on a first order approximation of $\eta$ around a reference policy $\pito$ for the local optimization of the parameters of $\pit$. The distance between both is approximately measured by
\begin{equation*}
\begin{array}{c}
\xi^{\theta_0}_t(\theta) :=	\frac{\pit(a_t|s_t)}{\pito(a_t|s_t)}.
\end{array}
\end{equation*}
The samples $(s_t, a_t)$ to be evaluated are generated within the computation of a rollout set $\M$. For this purpose a total of $N$ data points is computed by following the reference policy $\pito$. If a terminating state is reached in episodic tasks, a new start state is sampled with respect to $\Ps_0$. Altogether, the PPO objective for training the policy $\pit$ is given by
\begin{equation}
\begin{array}{l}
\label{eq:zeta_pi}
\zeta_\pi(\theta) := - \E_{(s_t, a_t) \in \M} \Big(\min \big[ \xi^{\theta_0}_t(\theta)\hat{A}_t, \\ \qquad\qquad\qquad\qquad\clip(\xi^{\theta_0}_t(\theta), 1-\varepsilon, 1+\varepsilon)\hat{A}_t\big] \Big)	
\end{array}
\end{equation}
which introduces a pessimistic balancing of two terms controlled by the $\clip$ parameter $\varepsilon > 0$ \cite{ppo}. The advantage of the $t^{\text{th}}$ data point is approximated by $\hat{A}_t$ which can further be used to define a cost function $\zeta_V$ for $V^\theta$ as the quadratic error
\begin{equation}
\label{eq:zeta_V}
\begin{array}{c}
\zeta_V(\theta) := - (\hat{A}_t)^2.
\end{array}
\end{equation}
Finally, a robust approximation $\hat{A}_t$ can be computed using the generalized advantage estimation \cite{gae} given as
\begin{equation*}
\begin{array}{c}
\hat{A}_t^{\gamma, \lambda} := \sum\limits_{k=0}^{\infty} (\gamma \lambda)^k \delta_{t+k}^{\theta},
\end{array}
\end{equation*}
which allows a sophisticated trade-off between bias and variance through the parameter $\lambda > 0$. Since the rollout set $\M$ is given by trajectories, this estimate can be computed for each data point by summing up to the end of the corresponding episode. Our implementation of the PPO method is summarized in Algorithm \ref{alg:ppo}.

\begin{algorithm}
	
	Initialize $\pit$ and $V^\theta$\;
	
	\While{not converged}
	{
		Set $\pito \leftarrow \pit$\;
		
		Generate a rollout set $\M$ following $\pito$ and compute $\hat{A}_t^{\gamma,\lambda}$ for each data point\;
		
		\For{K steps}
		{
			Draw a random batch of $M$ data points from $\M$\;
			Update $\pit$ and $V^\theta$ using a stochastic gradient descent algorithm with backpropagation and the cost functions $\zeta_\pi$ \eqref{eq:zeta_pi} and $\zeta_V$ \eqref{eq:zeta_V}\; 
		}	
		
	}
	
	\caption{Proximal policy optimization }
	\label{alg:ppo}
\end{algorithm}

\section{DEEP CONTROLLER FOR AUTONOMOUS DRIVING}
\label{sec:deep_controller}

\begin{figure}[t]
	\centering
	\begin{subfigure}{0.15\textwidth}
		\includegraphics[width=1.0\textwidth]{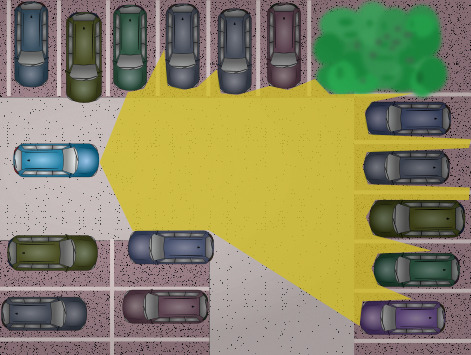}
		\subcaption{Sensors.}
		\label{fig:exploration_sensors}
	\end{subfigure}	
	\hfill
	\begin{subfigure}{0.15\textwidth}
		\includegraphics[width=1.0\textwidth]{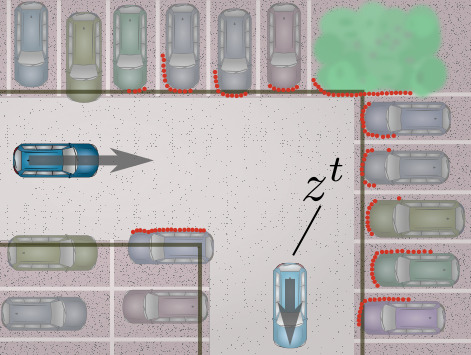}
		\subcaption{Target state $z^t$.}
		\label{fig:exploration_states}
	\end{subfigure}	
	\hfill
	\begin{subfigure}{0.15\textwidth}
		\includegraphics[width=1.0\textwidth]{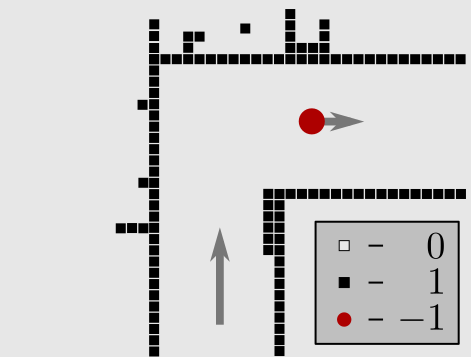}
		\subcaption{Perception $\O$.}
		\label{fig:exploration_perception}
	\end{subfigure}	
	\caption{During the exploration of a real parking lot, obstacles are perceived by sensors (a). Together with its knowledge about the drivable area, a current target state is defined (b). The surrounding from the perspective of the vehicle can be described by a coarse perception map where the target is represented by a red dot (c).}
	\label{fig:exploration}
\end{figure}

This work is part of the research project \textsc{AO-Car} whose objective is the development of algorithms for navigation and optimal control of autonomous vehicles in an urban environment. Here, information about the vehicle's surrounding are measured by, e.\,g., laser scanners and are further extended by a rough knowledge about the geometry of the drivable area (see \figref{fig:exploration}). Based on this, a desired target state $z^t$, including a speed value, is defined as illustrated in \figref{fig:exploration_states} for the exemplary situation of the autonomous parking lot exploration. The measurements and targets are updated at high frequency, ultimately resulting in a control loop. The task of the corresponding controller is to provide steering and acceleration values at every iteration, so that a safe trajectory to the target is obtained. Within this setting, all other vehicles are assumed to be static \cite{Sommer2018icatt}.

Training a deep reinforcement learning agent to implement a controller for solving this task involves the definitions of a corresponding MDP and the topologies of the neural networks $\pit$ and $V^\theta$ in accordance to it. Finally, an appropriate environment has to be designed in which the network parameters are learned through Algorithm~\ref{alg:ppo}. Here, in contrast to \cite{rl_driving_reality}, training is preformed within a simulation in order to quickly obtain a model as general as possible from various and diverse situations. Within the resulting episodic MDP, the policy has to make the vehicle reach the desired target state including a specified speed value. In particular, this should lead to a controlled stop, if the latter is zero. After this training step, the resulting agent can be deployed to control a real-world vehicle. Details of the simulated MDP are presented in the following.

\subsection{State and Action Spaces}

\begin{figure}[t]
	
	\centering
	\small
	\def\svgwidth{0.4\textwidth}
\begingroup%
  \makeatletter%
  \providecommand\color[2][]{%
    \errmessage{(Inkscape) Color is used for the text in Inkscape, but the package 'color.sty' is not loaded}%
    \renewcommand\color[2][]{}%
  }%
  \providecommand\transparent[1]{%
    \errmessage{(Inkscape) Transparency is used (non-zero) for the text in Inkscape, but the package 'transparent.sty' is not loaded}%
    \renewcommand\transparent[1]{}%
  }%
  \providecommand\rotatebox[2]{#2}%
  \newcommand*\fsize{\dimexpr\f@size pt\relax}%
  \newcommand*\lineheight[1]{\fontsize{\fsize}{#1\fsize}\selectfont}%
  \ifx\svgwidth\undefined%
    \setlength{\unitlength}{259.15065495bp}%
    \ifx\svgscale\undefined%
      \relax%
    \else%
      \setlength{\unitlength}{\unitlength * \real{\svgscale}}%
    \fi%
  \else%
    \setlength{\unitlength}{\svgwidth}%
  \fi%
  \global\let\svgwidth\undefined%
  \global\let\svgscale\undefined%
  \makeatother%
  \begin{picture}(1,0.46508679)%
    \lineheight{1}%
    \setlength\tabcolsep{0pt}%
    \put(0,0){\includegraphics[width=\unitlength,page=1]{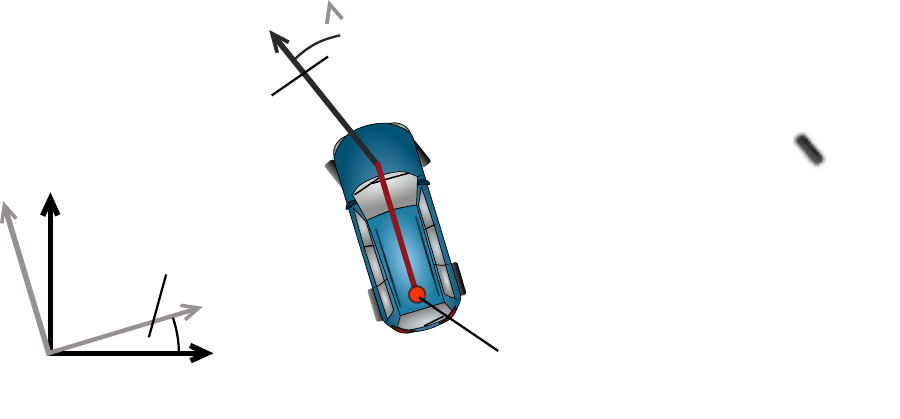}}%
    \put(0.5239061,0.01950524){\color[rgb]{0,0,0}\makebox(0,0)[lt]{\lineheight{0}\smash{\begin{tabular}[t]{l}$(x,\,y)$\end{tabular}}}}%
    \put(0.25920304,0.32274914){\color[rgb]{0,0,0}\makebox(0,0)[lt]{\lineheight{0}\smash{\begin{tabular}[t]{l}$\beta$\end{tabular}}}}%
    \put(0.3915843,0.43133176){\color[rgb]{0,0,0}\makebox(0,0)[lt]{\lineheight{0}\smash{\begin{tabular}[t]{l}$v$\end{tabular}}}}%
    \put(0.79427138,0.15462047){\color[rgb]{0,0,0}\makebox(0,0)[lt]{\lineheight{0}\smash{\begin{tabular}[t]{l}$L$\end{tabular}}}}%
    \put(0.04187711,0.27289851){\color[rgb]{0,0,0}\makebox(0,0)[lt]{\lineheight{0}\smash{\begin{tabular}[t]{l}$y$\end{tabular}}}}%
    \put(0.21912157,0.01134264){\color[rgb]{0,0,0}\makebox(0,0)[lt]{\lineheight{0}\smash{\begin{tabular}[t]{l}$x$\end{tabular}}}}%
    \put(0.17014175,0.18321655){\color[rgb]{0,0,0}\makebox(0,0)[lt]{\lineheight{0}\smash{\begin{tabular}[t]{l}$\varrho$\end{tabular}}}}%
    \put(0,0){\includegraphics[width=\unitlength,page=2]{einspurmodell_advanced.pdf}}%
  \end{picture}%
\endgroup%

	\caption{Left: Coordinates of the vehicle state vector. Right: Single track model simplification.}
	\label{fig:vehicle_coordinates}
\end{figure}

The research vehicle can be controlled algorithmically by specifying the steering wheel angle $\nu$ as well as the longitudinal acceleration $a$. The former can be mapped bijectively to the mean angle of the front wheels, defined as $\beta$. To prevent arbitrary fast changes, the angular velocity $\omega = \dot{\beta}$ is defined as part of the action space instead of $\beta$ or $\nu$, leading to
\begin{equation*}
\begin{array}{c}
\A := \{(a, \omega)^\top\in \R^2\;|\; \text{both bounded} \}.
\end{array}
\end{equation*}
These control variables have direct influence on the set of vehicle coordinates
\begin{equation*}
\begin{array}{c}
\Z := \{(x, y, v, \varrho, \beta)^\top \in \R ^5\;|\;v,\varrho,\beta \,\text{bounded}\}
\end{array}
\end{equation*}
as shown in \figref{fig:vehicle_coordinates}. The tuple $(x, y)$ describes the position of the center of the vehicle's rear axle with respect to an inertial system. The speed in the longitudinal direction is called $v$, where $\dot{v}=a$, and the orientation of the vehicle with respect to the inertial system is referred to as $\varrho$. 

Both, position and orientation of the current target $z^t\in \Z$, can be expressed by the bounded relative position $(x^r, y^r)$ with respect to the vehicle coordinates $z\in\Z$ and the complex number representation $(\varrho^r_\Re, \varrho^r_\Im)$ of the corresponding relative orientation. The form of the latter has the advantage of avoiding discontinuities. Moreover the controller needs to know the desired target speed as well as the current speed of the vehicle in order to predict its next states and to allow safe driving maneuvers. The target steering angle $\beta^t$ is not of interest, which ultimately leads to the first part of the state space
\begin{equation*}
\begin{array}{c}
\tilde{\S} = \{(x^r,y^r, v, v^t, \beta, \varrho^r_\Re, \varrho^r_\Im)^\top \in \R^7 \; | \; \text{all bounded} \}.
\end{array}
\end{equation*}

The second component is given by the vehicle's relative perception of its surrounding, combined with the a priori knowledge about the drivable area and the position of the target. This can be defined by a coarse grid $\O$ of size $n\times m$ with entries in $\{-1,0,1\}$, as shown in \figref{fig:exploration_perception}. Finally the MDP state space is specified by
\begin{equation*}
\begin{array}{c}
\S := \{(\tilde{s}, \O) \;|\; \tilde{s}\in \tilde{\S}, \O \in \{-1,0,1\}^{n\times m}\}.
\end{array}
\end{equation*}

\subsection{State Transitions}

The action-dependent transition between states of a real vehicle can be incorporated within the simulation using a system of differential equations to describe its behavior. This leads to an update of the relative vehicle coordinates as well as a new measurement of the obstacle perception. For low speeds, kinematic considerations such as simple single-track-models as in \cite{single_track_validation} or \cite{NonholonomicCar:Luca} are sufficient. Here, the vehicle is assumed to have only one wheel at the front and back respectively, each centered between the real ones as shown in \figref{fig:vehicle_coordinates}. 
This leads to the system of equations
\begin{equation*}
\begin{array}{c}
\begin{pmatrix}
\dot{x} \\
\dot{y} \\
\dot{v} \\
\dot{\varrho} \\
\dot{\beta}
\end{pmatrix} = \begin{pmatrix}
v\cos(\varrho) \\
v\sin(\varrho) \\
a \\
\nicefrac{v}{L} \tan(\beta) \\
\omega
\end{pmatrix},
\end{array}
\end{equation*}
where $L$ is the vehicle-specific wheelbase. Further physical constraints, such as a limited steering angle, can be directly considered within the simulation.

\subsection{Reward Function}

The reward function should encode the goal of the agent to reach the given target position with an appropriate orientation and speed. However, since the MDP at hand is continuous, it is almost impossible to ever fulfill this task starting with a random policy and to learn from such a success. This is in particular true in the difficult case of a target speed of zero, which requires an exact stopping maneuver at the end of the corresponding episode. 

As a result, training is performed in two phases and two separate policies, \textsc{driver} and \textsc{stopper}, are learned depending on the task. While the former is rewarded for quickly reaching the desired speed, the \textsc{stopper} should approach the target slowly and stop in the end. 
For both models, the first learning phase rewards proximity at every time step of the episode. Reaching the final position or rather performing a stop is especially highly rewarded. For the exemplary case of the \textsc{driver} this results in
\begin{align*}
	r_1^D := c_p\xoverline{\Delta p} + [r_0]_p + [\nicefrac{1}{2} r_0]_{p,\varrho}, \quad c_p, r_0 \in \R_+,
\end{align*}
where $\xoverline{\Delta p}$ measures the squared proximity to the target and $[\cdot]_p$ (or $[\cdot]_{p,\varrho}$) is only granted if the agent reaches the desired position (or position and orientation respectively). As soon as a policy learned to fulfill the goals of the first phase, additional behaviors such as an appropriate speed and small steering angles are taken into account, which is given by
\begin{align*}
	r_2^D := r_1^D + c_v\xoverline{\Delta v} + c_\beta\xoverline{\Delta \beta}, \quad c_v, c_\beta \in \R_+
\end{align*}
in case of the \textsc{driver}. Here, $\xoverline{\Delta v}$ and $\xoverline{\Delta \beta}$ measure the proximity of speed and current steering angle. For learning the \textsc{stopper} model, similar rewards apply and are complemented with, e.\,g., an additional weight on the desired speed in the first phase.

\subsection{Policy and Value Function as Neural Networks}

\begin{figure}[b]
	\centering
	\small
	\def\svgwidth{0.45\textwidth}
	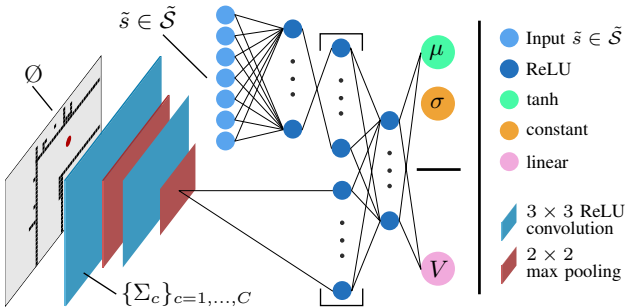
	\caption{Neural network topology of policy $\pit$ (top) and value function $V^\theta$ (bottom).}
	\label{fig:neural_networks}
\end{figure}

As suggested in \secref{sec:deep_reinforcement_learning}, two neural networks are trained to be identified as policy $\pit$ and value function $V^\theta$. 
Here, both share the same topology but not the same parameters.
The state component $\tilde{s} \in \tilde{\S}$
is processed by two dense layers of each $200$ ReLU activations, as shown in \figref{fig:neural_networks}. On the other hand, the evaluation of the perception map $\O$ is based on two convolutions, which allow to learn from the structure of the input. The spatial dimension is halved by a max pooling operation both times. While the first convolution consists of $C=30$ feature maps $\{\Sigma_c\}_{c=1,\ldots,C}$, the latter is reduced to only one, which is then flattened and also processed by a dense layer of $200$ ReLU activations. The result of both inputs is then concatenated and passed through a last $200$ ReLU layer. 

Finally, the output of the value function model $V^\theta$ is computed by a subsequent dense layer of one linear activation. Furthermore, the return of the policy $\pit$ is a pair $(\mu, \sigma)$ for each possible action, which is identified with the mean and standard deviation of a gaussian probability distribution. While the former is computed based on a $\tanh$ activation, the latter is defined to be independent from the input, which yields a general measure of how certain the model is about its actions. In particular, the noise introduced by $\sigma$ controls the policy's level of exploration when defining the rollout set~$\M$. Furthermore, it results in robustness against disturbances in the expected state transition with respect to the most preferred action $\mu$.

Constraints on the control commands can be incorporated by properly scaling the $\tanh$ activation and an additional $\clip$ operation in the case of action selection based on the normal distribution.

\section{TRAINING AND RESULTS}

\begin{figure}[b]
	\centering
	\includegraphics[width=0.11\textwidth]{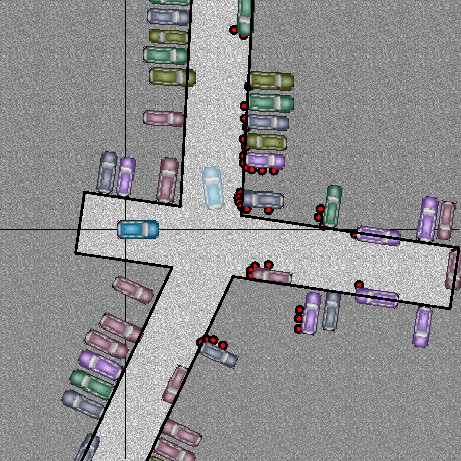}
	\includegraphics[width=0.11\textwidth]{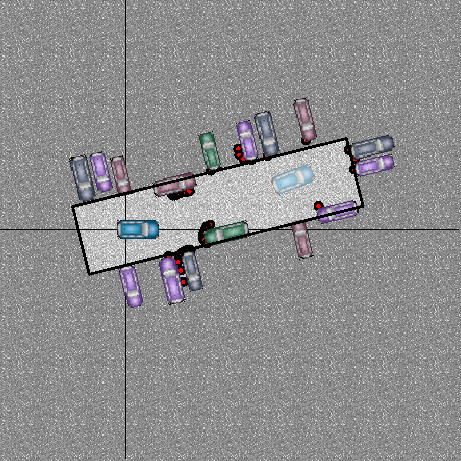}
	\includegraphics[width=0.11\textwidth]{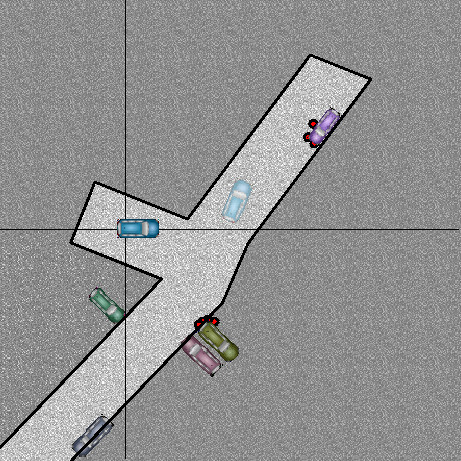}
	\includegraphics[width=0.11\textwidth]{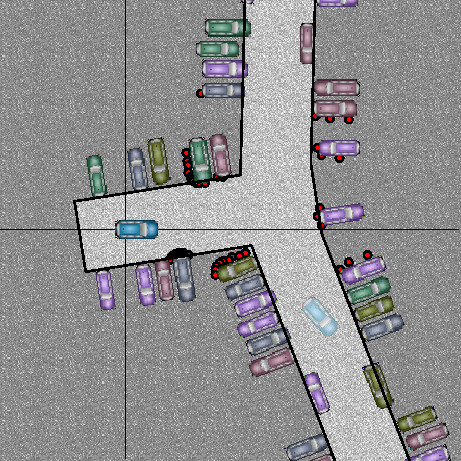}
	\\[1mm]
	\includegraphics[width=0.11\textwidth]{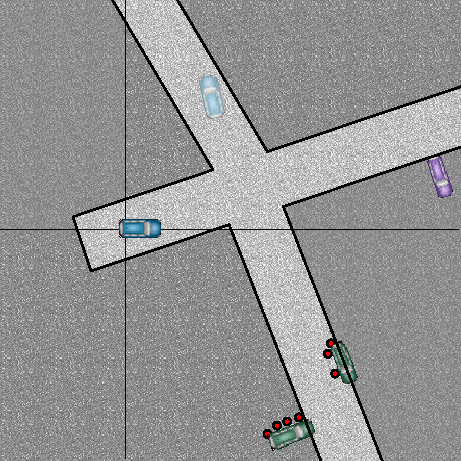}
	\includegraphics[width=0.11\textwidth]{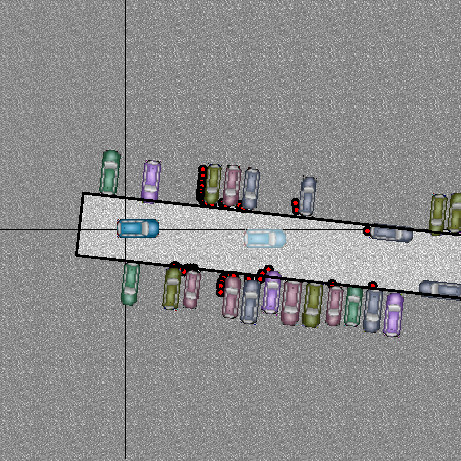}
	\includegraphics[width=0.11\textwidth]{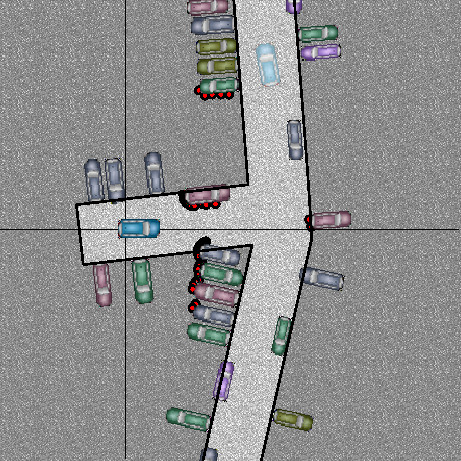}
	\includegraphics[width=0.11\textwidth]{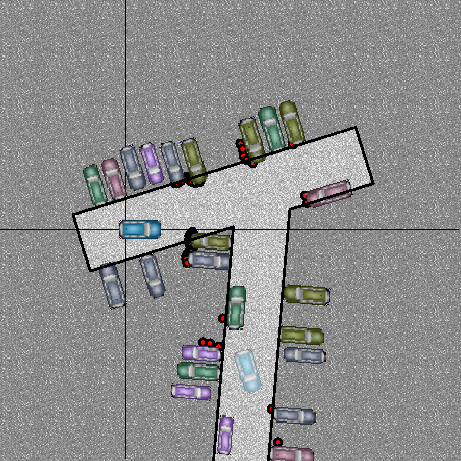}
	\caption{Randomly selected scenarios computed by the simulator. The initial state is represented by the dark blue vehicle at the center of the coordinate system and includes random values for speed $v$ and steering angle $\beta$. The light blue car defines the target. Red dots indicate sensor measurements of obstacle vehicles and the drivable area is defined by a closed polygon.}
	\label{fig:simulation}
\end{figure}

We evaluate our training procedure and present the performance of the resulting policy in simulation and on a real vehicle. Learning is realized on a GTX 1080 GPU while the simulator is conducted by an Intel Xeon E5 CPU kernel. Within the latter, random control tasks are generated as shown in \figref{fig:simulation}. The reward is shaped to make the policy drive the vehicle to the target state, which would define the end of an episode. Alternative termination criteria are the collision with an obstacle or the boundary polygon, exceeding a speed value of $3.3\,\nicefrac{\text{m}}{\text{s}}$ $\left(12\,\nicefrac{\text{km}}{\text{h}}\right)$ as well as reaching the maximum time step $T=250$. The simulated time between two such steps is defined to be $100\,\text{ms}$. The control values $a$ and $\omega$ are bounded by $\pm1.2\,\nicefrac{\text{m}}{\text{s}^2}$ and $\pm1.2\,\nicefrac{\text{rad}}{\text{s}}$ respectively.

Proximal policy optimization is carried out by collecting rollout sets consisting of $16{,}384$ time steps, which are then used within $K=16$ optimization steps with batches of size $M=1{,}024$. One such epoch takes on average $150\,\text{s}$, while a maximum of $350$ epochs is required for convergence. This results in a training time of less than $15\,\text{h}$. However, policies of similar quality can be obtained even without incorporating other vehicles as obstacles during the learning process, as outlined in the next section. The resulting simulation is much faster without detailed computation of the agent's perception, which leads to optimization cycles of $90\,\text{s}$. This reduces the maximum training time to $9\,\text{h}$.

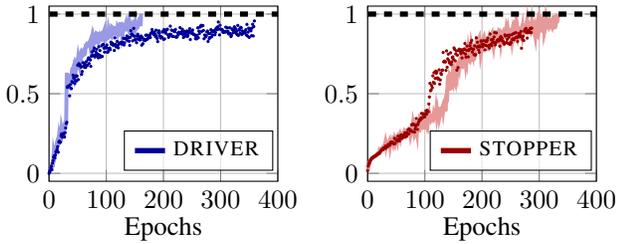
\begin{figure}[t]
	\centering
	
	\pgfplotsset{
		compat=newest,
		/pgfplots/legend image code/.code={%
			\draw[#1] 
			plot coordinates {
				(0cm,0cm) 
				(0.4cm,0cm)
			};
		},
	}
	
	\begin{tikzpicture}
	\begin{axis}[
	xmin = 0, 
	xmax = 400,
	ymin = -0.05,
	ymax = 1.05,	   		
	xlabel = Epochs,
	legend style={at={(0.98,0.06)},anchor=south east, fill opacity=0.6, text opacity=1}, 
	grid = both,
	width=0.26\textwidth,
	height=0.22\textwidth,
	x label style={at={(axis description cs:0.5,-0.15)},anchor=north},]  	   		
	
	\addplot[no marks, line width = 2.0, solid, color=b1] coordinates {
		(1, 2)
		(400, 2)
	};
	\addlegendentry{\textsc{driver}}
	
	\addplot[no marks, line width = 1.5, color=b2, smooth] table [x expr = \coordindex, y expr=(\thisrow{reward}-2.0)/(141-2.0), col sep=comma] {models/test36_DN_training.log};	   		
	
	\addplot[only marks, mark size = 0.4, color=b1] table [x expr = \coordindex, y expr=(\thisrow{reward}-2.0)/(141.5-2.0), col sep=comma] {models/driver_test39_training.log};
	
	\addplot[no marks, line width = 2, dashed] coordinates {
		(1, 1)
		(400, 1)
	};
	
	\end{axis}
	\end{tikzpicture}   		
	\begin{tikzpicture}
	\begin{axis}[
	xmin = 0, 
	xmax = 400,
	ymin = -0.05,
	ymax = 1.05,
	xlabel = Epochs,
	legend style={at={(0.98,0.06)},anchor=south east, fill opacity=0.6, text opacity=1}, 
	grid = both,
	width=0.26\textwidth,
	height=0.22\textwidth,
	x label style={at={(axis description cs:0.5,-0.15)},anchor=north},]

	\addplot[no marks, line width = 2.0, solid, color=r1] coordinates {
		(1, 2)
		(400, 2)
	};
	\addlegendentry{\textsc{stopper}}
	
	\addplot[no marks, line width = 1.5, color=r2, smooth] table [x expr = \coordindex, y expr=(\thisrow{reward}+31.5)/(211.5+31.5), col sep=comma] {models/test62_SN_training.log}; 			
	
	\addplot[only marks, mark size = 0.4, color=r1] table [x expr = \coordindex, y expr=(\thisrow{reward}+31.5)/(211.5+31.5), col sep=comma] {models/stopper_test64_training.log};   	
	
	\addplot[no marks, line width = 2, dashed] coordinates {
		(1, 1)
		(400, 1)
	};

	\end{axis}
	\end{tikzpicture}

	\caption{Normalized average reward during learning of the \textsc{driver} and \textsc{stopper} policy. The dotted lines represent training with obstacle vehicles (type A), while the solid lines indicates that they were excluded (type B). Progress is displayed until the respective maximum value is reached.}
	\label{fig:training_reward}
\end{figure}

The typical development of the average reward during training is shown in \figref{fig:training_reward}. As one would suggest, convergence of a \textsc{stopper} policy is much slower in the beginning, due to the fact that it is harder for the agent to learn about its final goal. \textsc{Driver} models usually converge in less than $200$ epochs in the simplified setting, which results in a computation time of $5\,\text{h}$. Furthermore, the overall rewards are considerably lower in the case of obstacle vehicles included during learning. Since that task is more difficult, this result is also to be expected.

\subsection{Results within the Simulation}

We evaluate \textsc{driver} and \textsc{stopper} models resulting from learning with obstacle vehicles (type A) and without them (type B), while all other parameters remain identical. In particular, we demonstrate their performance within the simulated environment which is summarized in table \ref{tab:performance}. The evaluation is done based on $10{,}000$ randomly generated control tasks including other vehicles. Results show high success rates with other terminations mainly due to obstacles. Further investigations show that this outcome is very often caused by an infeasible combination of initial orientation, speed and steering angle with respect to the placement of obstacles which would trigger an emergency brake in reality. Even though type B models were only trained based on the polygon representing the drivable area, they still achieve results of similar quality than the type A agent. In case of the \textsc{driver} model they are even better. Most importantly this indicates that type B models are capable of handling unknown obstacles, even if those lead to constrictions on the lane.

\begin{table}[b]
	\centering
	\caption{Comparison of termination criteria and average reward of models trained with (type A) and without obstacles (type B). Values are generated in $10{,}000$ simulation runs including other vehicles. The average cumulated reward per episode is normalized to their respective maximum.}
	\label{tab:performance}
	\begin{tabular}{c|cc!{\vrule width 1pt}c|c|c|c|c}
		& \multicolumn{2}{c!{\vrule width 1pt}}{\textbf{Obstacle}}  & & \multicolumn{4}{c}{\textbf{Termination [\%]}} \\
		\textbf{Policy} & \multicolumn{2}{c!{\vrule width 1pt}}{\textbf{Training}} & {\boldmath$\varnothing$} \textbf{Rew.} & \textbf{Succ.} & \textbf{Coll.} & \textbf{Time} & \textbf{Speed}\\
		\Xhline{1pt}
		\multirow{2}{*}{\textsc{Driver}} & \checkmark & (A) & 1.0 & 90.7 & 6.1 & 3.2 & 0.0 \\
		& - & (B) & 0.984 & 87.5 & 10.5 & 2.0 & 0.0 \\
		\hline
		\multirow{2}{*}{\textsc{Stopper}} & \checkmark & (A) & 0.967 & 80.8 & 11.5 & 7.8 & 0.0 \\
		& - & (B) & 1.0 & 82.9 & 14.5 & 2.6 & 0.0
	\end{tabular}	
\end{table}

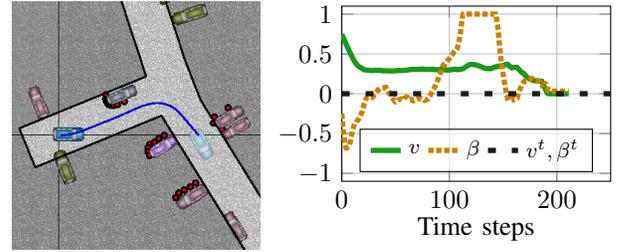
\begin{figure}[t]
	
	\pgfplotsset{
		compat=newest,
		/pgfplots/legend image code/.code={%
			\draw[#1] 
			plot coordinates {
				(0cm,0cm) 
				(0.4cm,0cm)
			};
		},
	}
	
	\centering

	\includegraphics[width=0.185\textwidth]{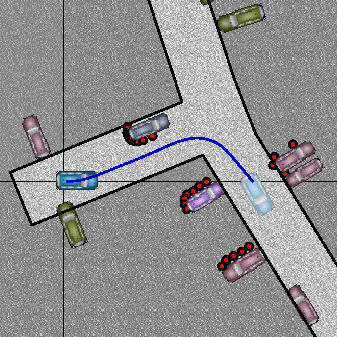}
	\begin{tikzpicture}
	\begin{axis}[
	xmin = 0, 
	xmax = 250,
	ymin = -1.1,
	ymax = 1.1,
	xlabel = Time steps,
	legend style={at={(0.5,0.3)},anchor=north, font=\footnotesize, fill opacity=0.6, draw opacity=1, text opacity=1, legend columns=-1}, 
	grid = both,
	width=0.29\textwidth,
	height=0.22\textwidth,
	x label style={at={(axis description cs:0.5,-0.15)},anchor=north},]
	
	\addplot[no marks, line width = 2, color=g2] table [x expr = \coordindex, y expr=\thisrow{v}/(14/3.6), col sep=comma] {solution/simple/full_states.csv};
	\addlegendentry{$v$}
	
	\addplot[no marks, line width = 2, color=o1, densely dotted] table [x expr = \coordindex, y expr=\thisrow{beta}/(0.55), col sep=comma] {solution/simple/full_states.csv};
	\addlegendentry{$\beta$}
	
	\addplot[no marks, line width = 2, loosely dashed, color=k1] coordinates {
		(0, 0)
		(250, 0)
	};
	\addlegendentry{$v^t, \beta^t$}
	
	
	%
	%
	%
	%

	\end{axis}
	\end{tikzpicture}
	\caption{Exemplary solution of a \textsc{stopper} agent which was trained without obstacle vehicles (type B). The values in the right plot are normalized to their respective maximum.}
	\label{fig:solution_trajectory}
\end{figure}

\begin{figure}[t]
	\centering
	\includegraphics[width=0.23\textwidth]{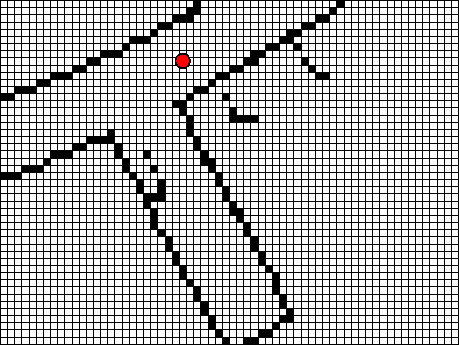}
	\hfill
	\footnotesize
	\def\svgwidth{0.23\textwidth}
	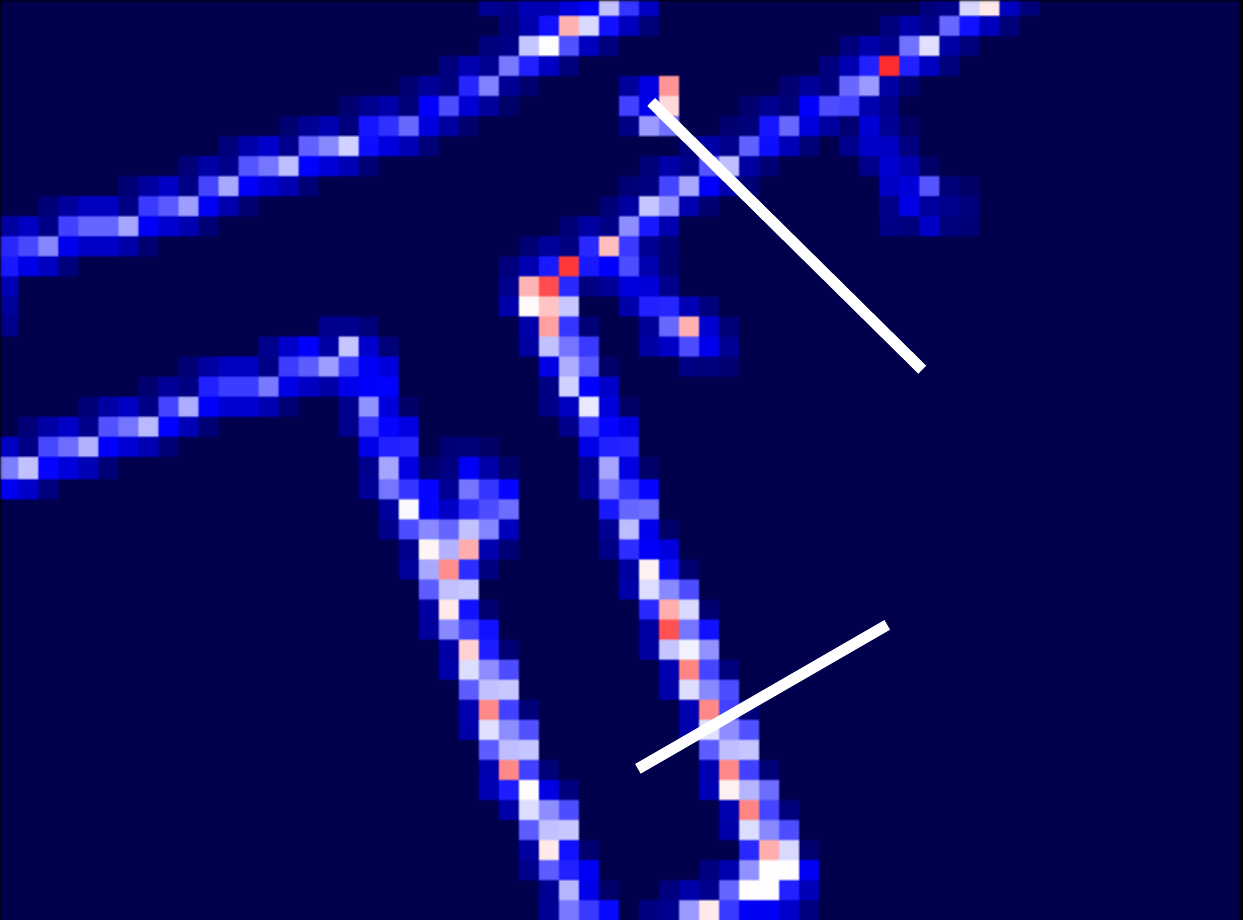
	\caption{Perception input $\O$ (left) and its attention map $\mathbb{A}$ (right) of the initial situation in \figref{fig:solution_trajectory}. While dark blue pixels indicate lowest attention, it is at maximum for red ones.}
	\label{fig:solution_perception}
\end{figure}

This result is further supported by \figref{fig:solution_trajectory} and \figref{fig:solution_perception} which present an exemplary task solved by a type B \textsc{stopper} policy. The former displays the path as well as the corresponding normalized speed and steering values when applying this agent. In particular, the agent is not pulling out before turning to take the obstacle vehicle on the lane into account. \figref{fig:solution_perception} shows the perception map $\O$ of the initial situation in \figref{fig:solution_trajectory} relative to the agent. To evaluate what parts of it are most important to the neural network, an \emph{attention map} $\mathbb{A}$ can be computed as suggested in \cite{attention}. For that the feature maps $\{\Sigma_c\}_{c=1,\ldots,C}$ of the first convolutional filter (see \figref{fig:neural_networks}) are squared and summed up, yielding $\mathbb{A} := \sum_{c=1}^{C}\Sigma_c^2.$ In the situation at hand, high attention is paid to the agent's immediate surrounding as well as to important points in the long term such as the corner, where it has to turn around, or the surrounding of the target. In addition, comparably high attention is paid to the obstacle standing on the lane in front of it. Altogether, these results show that a high quality agent can be obtained even with a simple simulation in short training time.

\subsection{Results on a Real Vehicle}
After the training in simulation has been completed, a \textsc{stopper} and \textsc{driver} model are combined into one deep controller. This is applied within the system for autonomous driving, developed as part of the research project \mbox{\textsc{AO-Car} \cite{Sommer2018icatt}}, as the main controller during the autonomous exploration of a parking lot (as described in \secref{sec:deep_controller}). 
Experiments are performed on a standard Volkswagen Passat GTE Plug-in-Hybrid with additional laser scanners at its front and rear\footnote[2]{\label{note:aocar}For further details visit \url{www.math.uni-bremen.de/zetem/aocar}}. 

During exploration, an estimate of the vehicle's state is computed every $20\,\text{ms}$ using an Extended Kalman Filter as described in \cite{clemens2016ekf}. From this, a current target state is deduced and corresponding control commands are computed by the controller. Due to the target being updated at high frequency, the vehicle continues driving until a position where to stop is provided. Since the outputs of the deep controller are the acceleration and the steering angle velocity, the latter is numerically integrated for another $20\,\text{ms}$ to define the desired steering angle and thus ultimately the vehicle control. The computation of one control command by the neural network representing the policy takes approximately $1.2\,\text{ms}$ on an Intel Core i7-4790 CPU.

\begin{figure}[t]	
	\begin{minipage}{0.45\textwidth}
		\centering
		\begin{tikzpicture}
		\node (img)  {\includegraphics[scale=0.6]{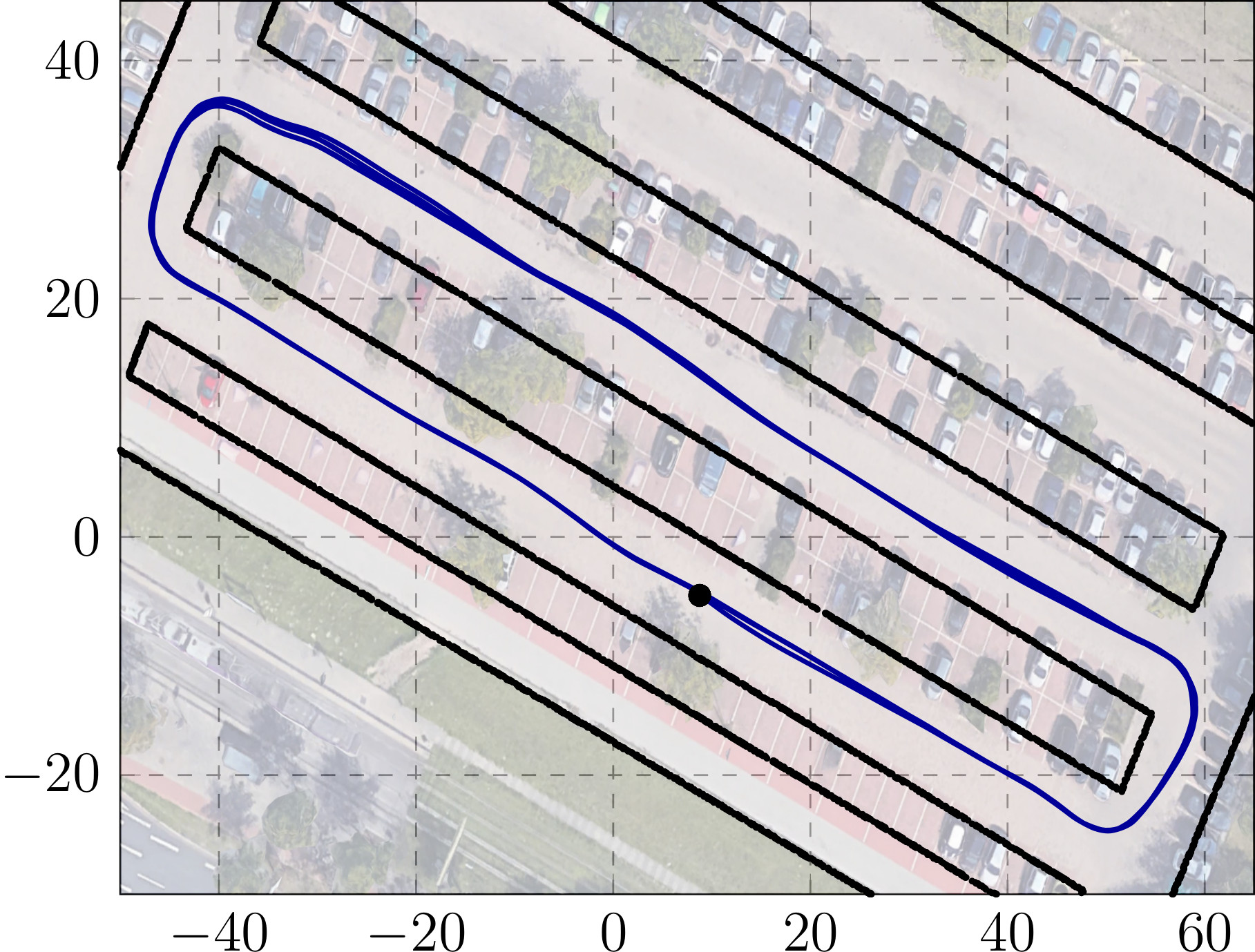}};
		\node[below=of img, node distance=0cm, yshift=1cm, xshift=0.3cm] {relative $x$ position};
		\node[left=of img, node distance=0cm, rotate=90, anchor=center,yshift=-0.8cm, xshift=0.1cm] {relative $y$ position};
		\end{tikzpicture}
	\end{minipage}%
	\caption{Exemplary path (blue line) taken by the deep controller when five times circling a real parking area counterclockwise. The positions are relative to the prior knowledge about the parking lot (black lines). The start is marked by a black dot. The underlying satellite image (\textsc{{\textcopyright} 2009 GeoBasis-DE/BKG}) is inserted as a reference only (so it does not show the actual occupancy of parking spaces).}
	\label{fig:testdrive_path}
\end{figure}

General characteristics of the deep controller can be evaluated on the basis of \figref{fig:testdrive_path}. The driven path is in the center of the lane at all times, leading to a high safety of the corresponding trajectory. This applies in particular to the turning maneuvers, after which the vehicle is immediately aligned with the lane again. One can further notice that the paths taken in every turn are very similar to each other. The only deviations occur at the start or in the case of a third party vehicle influencing the control commands (top left). 

\begin{figure}[t]
	\centering
	\includegraphics[height=0.16\textwidth]{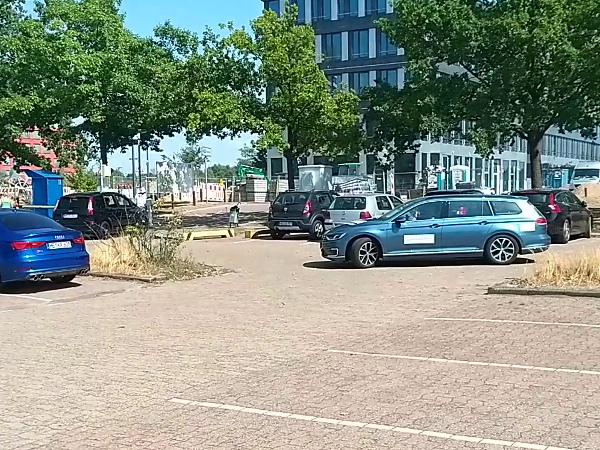}
	\hfill
	\setlength{\fboxsep}{0pt}%
	\setlength{\fboxrule}{0.1mm}%
	\fbox{\includegraphics[height=0.16\textwidth]{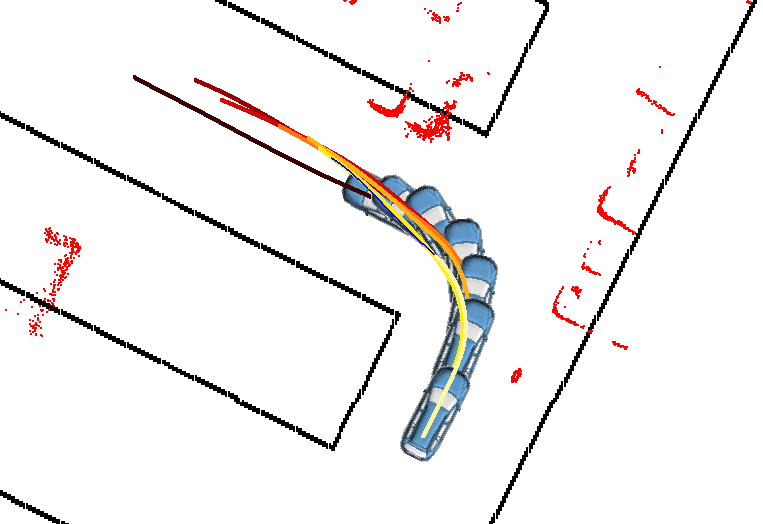}}
	\\[1mm]
	\includegraphics[height=0.16\textwidth]{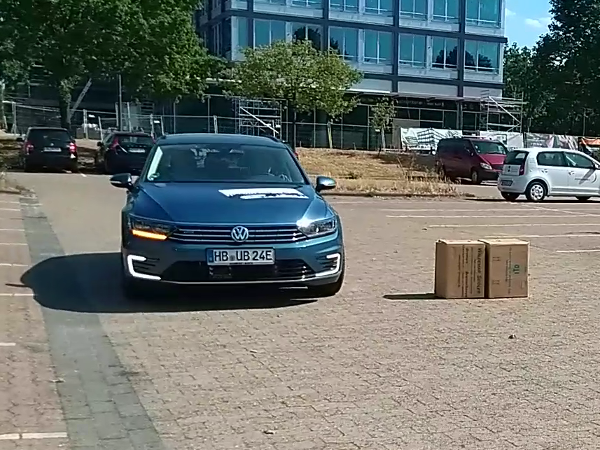}
	\hfill
	\setlength{\fboxsep}{0pt}%
	\setlength{\fboxrule}{0.1mm}%
	\fbox{\includegraphics[height=0.16\textwidth]{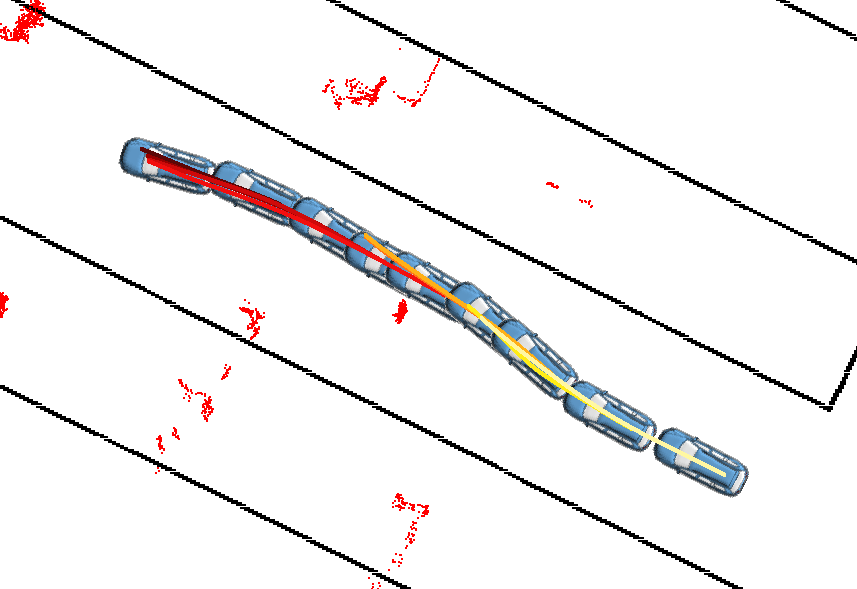}}
	\\[1mm]
	\includegraphics[height=0.16\textwidth]{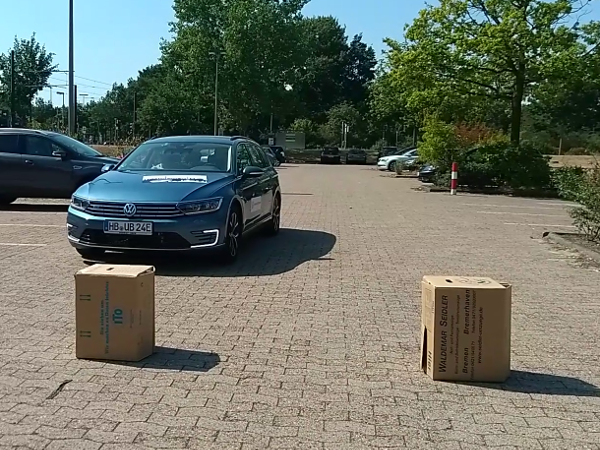}
	\hfill
	\setlength{\fboxsep}{0pt}%
	\setlength{\fboxrule}{0.1mm}%
	\fbox{\includegraphics[height=0.16\textwidth]{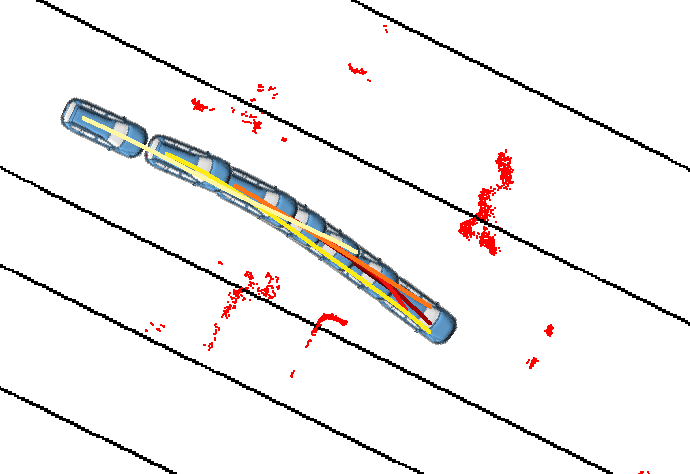}}
	\caption{Application of the deep controller on a real vehicle in three exemplary scenarios. Red dots indicate laser measurements and the drivable area is marked by black lines. The currently planned path of the controller is given by trajectories with a color gradient from light yellow to dark red. Top: Turning. Middle: Driving around an obstacle and stopping afterwards. Bottom: Detection of a blockade on the street and stopping in front of it.}
	\label{fig:exploration_real}
\end{figure}

\figref{fig:exploration_real} shows three specific scenarios which can successfully and safely be handled by the proposed controller. In particular, we overlaid a series of state estimates and the corresponding trajectories that the agent would execute based on the vehicle's current perception. One can see that the controller is able to perform highly challenging maneuvers such as sharp turnings with unknown obstacles. The system is further able to make decisions based on new objects which makes the controller execute an evasive maneuver or a soft stop. In all cases, following the deep controller leads to very smooth driving behavior of the research vehicle. Altogether, the presented method is able to provide sophisticated control commands while still being able to fulfill strict requirements on the computation time.

\section{CONCLUSION}

We presented a general design approach for a nonlinear controller that is able to provide highly advanced control commands in extremely short computation time. This was done by approximating the control problem within a simulation as a Markov decision process which was then solved by an agent in the setting of a reinforcement learning problem. Training was performed by a proximal policy optimization method with the policy being defined as a neural network. We evaluated our approach in the context of autonomous driving and showed that a high quality controller could be obtained within a few hours of training. Furthermore, we demonstrated its performance on a full-size research vehicle during the autonomous exploration of a parking lot. For example, the controller was able to handle sharp turnings as well as unknown obstacles by performing an evasive maneuver or a stop. In particular, this work is one of the first successful and, to the best of our knowledge, the currently most general application of a deep reinforcement learning agent to a real autonomous vehicle. Future work will include a more detailed analysis of the neural network structure and the state representation as well as applications to further scenarios. The training process as well as the evaluation on the research vehicle are available as a video at \\\url{https://youtu.be/1HwHdL7bY3A}.

\section*{APPENDIX}
\begin{table}[H]
	\centering
	\def\arraystretch{1.1}
	\caption{Overview of hyperparameters}
	\label{tab:params}
	\begin{tabular}{ p{.11\textwidth}|p{.02\textwidth}|p{.04\textwidth}||p{.06\textwidth}|p{.02\textwidth}|p{.04\textwidth} }
		\multicolumn{3}{c||}{\textsc{Training}} 													& \multicolumn{3}{c}{\textsc{Reward}}\\
		\hline
		\textbf{Descript.} 						& \textbf{Var.}				& \textbf{Value} 		& \textbf{Descript.} 			& \textbf{Var.}				& \textbf{Value} 			\\
		\hline				
		Steps / episode 						& $T$  						& $250$ 	 			& \textsc{Driver} 						& $c_p$ 					& $0.1$	\\
		Step length								&							& $0.1\,[\text{s}]$  	&  								& $c_v$ 					& $0.5$ \\
		Rollout size					 		& $N$						& $16{,}384$ 			&  								& $c_\beta$ 				& $0.5$ \\
		Optim. / rollout		 				& $K$ 						& $16$ 					&  								& $r_0$ 					& $50$ \\
		Minibatch size							& $M$ 						& $1{,}024$ 			& \multicolumn{3}{c}{} \\
		Discount 								& $\gamma$					& $0.99$  				& \multicolumn{3}{c}{} \\
		Gener. advantage						& $\lambda$	 				& $0.95$ 				& \multicolumn{3}{c}{\textsc{Dynamics}}\\\cline{4-6}
		$\clip$ parameter 						& $\varepsilon$ 			& $0.1$ 				& \textbf{Descript.} 			& \multicolumn{2}{l}{\textbf{Bounds}}\\\cline{4-6}
		Scaling of $\zeta_V$ 					&							& $0.1$ 				& Acc. 							& \multicolumn{2}{l}{$|a| \leq 1.2\,[\nicefrac{\text{m}}{\text{s}^2}]$}\\	
		Adam optimizer							& $\alpha$ 					& $5\mathrm{e}{-5}$ 	& Ang. vel. 					& \multicolumn{2}{l}{$|w| \leq 1.2\,[\nicefrac{\text{rad}}{\text{s}}]$}\\
		(cf. \cite{adam})						& $\beta_1$  				& $0.9$  				& Steering 						& \multicolumn{2}{l}{$|\beta| \leq 0.55\,[\text{rad}]$} \\
												& $\beta_2$ 				& $0.999$ 				& Speed 						& \multicolumn{2}{l}{$v \in [0, 3.3]\,[\nicefrac{\text{m}}{\text{s}}]$} \\
												& $\epsilon$				& $1\mathrm{e}{-5}$ 	& \multicolumn{3}{c}{} \\		
	\end{tabular}
\end{table}


\bibliographystyle{IEEEtran}
\bibliography{IEEEabrv,references}

\end{document}